# Syntactic Autonomy

## Or Why There is no Autonomy Without Symbols and how Self-Organizing Systems Might Evolve Them

**LUIS MATEUS ROCHA**
*Complex Systems Modeling Team, Computer Research and Applications Group*
*Los Alamos National Laboratory, MS B265, Los Alamos, NM 87545, USA*
e-mail: rocha@lanl.gov or rocha@santafe.edu URL: http://www.c3.lanl.gov/~rocha

**ABSTRACT**
Two different types of agency are discussed based on dynamically coherent and incoherent couplings with an environment respectively. I propose that until a private syntax (syntactic autonomy) is discovered by dynamically coherent agents, there are no significant or interesting types of closure or autonomy. When syntactic autonomy is established, then, because of a process of description-based selected self-organization, open-ended evolution is enabled. At this stage, agents depend, in addition to dynamics, on localized, symbolic memory, thus adding a level of dynamical incoherence to their interaction with the environment. Furthermore, it is the appearance of syntactic autonomy which enables much more interesting types of closures amongst agents which share the same syntax. To investigate how we can study the emergence of syntax from dynamical systems, experiments with cellular automata leading to emergent computation to solve non-trivial tasks are discussed. RNA editing is also mentioned as a process that may have been used to obtain a primordial biological code necessary open-ended evolution.

## 1. CLOSURE AND AGENTS

To understand *closure* in evolutionary systems, we need to be able to distinguish between an evolving system and its environment, which typically includes other such systems. Indeed, the notion of closure can only be meaningful if we can discern some sort of autonomy of one system from an environment or alternatively, a loop, or cycle, established between more than one system[i]. For both of these cases, in the context of evolutionary systems, we need first to understand the nature of distinguishable systems capable of evolving. In other words, we need to agree on what is it that allows us to separate a system from an environment so that we can refer to it as autonomous or place it in some kind of loop with other systems or the environment[ii]? This separability implies some kind of agency, so let us start by discussing the notion of *agent*.

The term agent is used today to mean anything between a mere subroutine to a conscious entity. There are "helper" agents for web retrieval and computer maintenance, robotic agents to venture into inhospitable environments, agents in an economy, etc. Intuitively, for an object to be referred to as an agent it must possess some degree of autonomy, that is, it must be in some sense distinguishable from its environment by some kind of spatial, temporal, or functional boundary. It must possess some kind of identity to be identifiable in its environment. To make the definition of agent useful, we often further require that agents must have some autonomy of action, that they can engage in tasks in an environment without direct external control. This leads us to an important definition of an agent from the XIII century, due to Thomas Aquinas[3]: an entity capable of election, or choice.

This is a very important definition indeed; for an entity to be referred to as an agent, it must be able to step out of the dynamics of an environment, and make a decision about what action to take next – a decision that may even go against the natural course of its environment. Since choice is a term loaded with many connotations from theology, philosophy, cognitive science, and so forth, I prefer to discuss instead the ability of some agents to step out of the dynamics of its interaction with an environment and explore different behavior alternatives. In physics we refer to such a process as *dynamical incoherence*[4]. In computer science, Von Neumann, based on the work of Turing on universal computing devices, referred to these systems as *memory-based systems*. That is, systems capable of engaging with their environments beyond concurrent state-determined interaction by using memory to store descriptions and representations of their environments. Such agents are dynamically incoherent in the sense that their next state or action is not solely dependent on the previous state, but also on some (random-access) stable memory that keeps the same value until it is accessed and does not change with the dynamics of the environment-agent interaction[iii]. In contrast, state-determined systems are dynamically coherent (or coupled) to their environments because they function by reaction to present input and state using some iterative mapping in a state space.

Let us then refer to the view of agency as a dynamically incoherent system-environment engagement or coupling as the *strong sense of agency*, and to the view of agency as some degree of identity and autonomy in dynamically

---

[i] Joslyn[1] presents a discussion of the types of closures one can identify.

[ii] For a discussion of possible types of system/environment couplings, particularly, of evolving couplings, please refer to [2].

[iii] Stable memory must of course be materially implemented, and thus it is always dynamic at some level[5]. The stability or inertness of memory tokens must be seen as relative stability within the timescale of the dynamic system-environment coupling. The inertness of memory tokens has been discussed in detail elsewhere[6, 7, 8].

coherent system-environment coupling as the *weak sense of agency*. The strong sense of agency is more precise because of its explicit requirement for memory and ability to effectively explore and select alternatives. Indeed, the weak sense of agency is much more subjective since the definition of autonomy, a boundary, or identity (in a loop) are largely arbitrary in dynamically coherent couplings, as we shall discuss below. I further posit that only the existence of symbolic memory tokens can establish clearly identifiable autonomies and closures, which in turn lead to open-ended evolution.

## 2. DYNAMICALLY COHERENT SYSTEM-ENVIRONMENT COUPLINGS: SITUATED SEMIOSIS

### 2.1 Self-Organization

Self-organization is seen as the process by which systems of many components tend to reach a particular state, a set of cycling states, or a small volume of their state space (attractor basins), with no external interference. This attractor behavior is often recognized at a different level of observation as the spontaneous formation of well organized structures, patterns, or behaviors, from random initial conditions (emergent behavior). The systems used to study this behavior computationally are referred to as discrete dynamical systems or state-determined systems, since their current state depends only on their previous state. They possess a large number of components or variables, and thus high-dimensional state spaces.

Computational self-organization is often used to model physical matter using such constructs as boolean networks or cellular automata. The state-determined transition rules are interpreted as the laws of some physical or chemical system[2]. It follows from the observed attractor behavior that there is a propensity for matter to self-organize[9]. In this sense, matter is described by the laws of physics (which are modeled computationally) and the emergent characteristics of self-organization. In the following, whenever the words matter and materiality are used, they should be understood as reflecting this notion of self-organization both in physical and computational environments.

### 2.2 Emergent Classification: Semantic Emergence

Self-organizing attractor basins can be used to refer to observables accessible to the self-organizing system in its environment, and thus perform environmental *classifications* (e.g. classifying neural networks). These self-organizing systems produce dynamic stabilities in interaction with their environments, which route observables from this interaction into a small set of attractor basins which produce a corresponding small set of behaviors. In this sense we say that the self-organizing system classifies its environment into such a small subset of behaviors. This process of obtaining classifications of an environment by a self-organizing system, has been referred to generally as *emergent classification*[6,7]. Emergent because it is the result of the local, state-determined, interaction of the basic components of the self-organizing system and its dynamic interaction with the environment. In this sense, the dynamically coherent coupling of a self-organizing system with its environment establishes a semantic dimension: attractor states leading to a small set of behaviors produced by the self-organizing system are used to classify and cope with an environment. Thus, the attractor states refer to environmental observables. A discussion of this emergent classification, also known as eigen-behavior, was pursued in [6].

## 2.3 Pragmatics: Selected Self-Organization

To effectively deal with a changing environment, systems capable of relating internal stabilities to environmental regularities, must be able to change their own dynamics in order to produce new basins of attraction for new classifications and behaviors. In other words, the self-organizing system must be structurally coupled to some external system which acts on the structure of the first inducing some form of explicit or implicit selection of its dynamic classifications, this process has been referred to as *selected self-organization* [6,7,8]. Now, for selection to occur we must have some internal vehicle for classification — there must be different alternatives. The range of attractor landscapes of self-organizing systems offers these alternatives. One way of conceptualizing this is to think of the attractor landscape as a distributed memory bank [10], where each attractor basin is seen as storing a given classification configuration. This ability of a self-organizing system to select appropriate dynamic classification configurations to deal with a changing environment can be referred to as *dynamically coherent selected self-organization* or *dynamically coherent categorization* (in the context of cognitive science[11]).

In the biological realm, such selection is implicitly defined by different rates of reproduction of individuals in varying (genetic) populations, while in the cognitive realm we may have some form of explicit selection referred to as learning through conversation[11]. A simple example in an applied domain, would be an external algorithm for selecting the weights (structural perturbation) of a neural network in order to achieve some desired classification, or evolutionary strategies which rely on internal random variation which is ultimately externally selected.

Selected self-organization explicitly emphasizes a second dimension of the semiosis of self-organizing systems dynamically coupled (in situation) with their environments. If self-organizing classification implies semantic emergence, selection implies pragmatic environmental influence. In fact, these two dimensions of semiosis cannot be separated; the meaning of the classifications of a self-organizing system does not make sense until it is grounded in the feedback from the repercussions it triggers in its environment. The dynamically coherent coupling, or situation, of a classifying, self-organizing agent in its environment is the source of meaning. Indeed, selection does not act on memory tokens internal to a classifying system but on the repercussions those trigger in an environment. Semantics of dynamically coherent (situated) agents is pragmatic. In this sense, meaning is not private to the agent but can only be understood in the context of the agent's situation in an environment with its specific selective pressures.

The ability of agent/environment couplings to select appropriate attractor states to cope with an environment takes us closer to agent's that can select behavior alternatives. Note, however, that selected self-organization requires both self-organization <u>and</u> a selective process to be specified[iv]. Not merely state-determined, rule-following, self-organization which would amount to agents with no real alternatives. This notion of selected self-organization leads us now to think of what kinds of selection processes are possible, and more importantly for the present work, does this choice of alternative classifications/behaviors exist in or require some kind of autonomy or closure?

---

[iv] The need of a serious study of selection processes has been outlined in [12], a more extensive discussion is left for future work.

## 3. DYNAMICALLY INCOHERENT SYSTEM-ENVIRONMENT COUPLINGS: OPEN-ENDED SEMIOSIS

### 3.1 Von Neumann and the Syntactic Advantage

Von Neumann's model of self-replication[13] is a systems-theoretic criteria for open-ended evolution[v]. Based on the notion of universal construction and description it provides a threshold of complexity after which systems that observe it can for ever more increase in complexity (open-ended evolution). However, unlike the situated semiosis of agents dynamically coherent with their environments described in 2, this model clearly does not rely on a distributed but on a local kind of memory. Von Neumann's descriptions entail a symbol system on which construction commands are cast. These descriptions are not distributed over patterns of activation of the components of a self-organizing system, but are instead localized on "inert" structures which can be used at any time — a sort of random access memory.

By "inert" structures, I mean components with many dynamically equivalent states which can be used to set up an arbitrary semantic relation with the environment. For instance, in the genetic system (which Von Neumann's model conceptually describes), most any sequence of nucleotides is equally possible, and its informational value (genetic information) is largely independent of the particular dynamic behavior of the DNA/RNA sequence. Genetic information is not expressed by the dynamics of nucleotide sequences (RNA or DNA molecules), but is instead mediated through an arbitrary coding relation that translates nucleotide sequences into amino-acid sequences whose dynamic characteristics ultimately express genetic information in an environment. It is precisely the dynamic irrelevance of nucleotide sequences ("inertness") that makes DNA/RNA ideal candidates for localized carriers of genetic information (descriptions) given an arbitrary genetic code[14, 15]. DNA qua carrier of genetic information in biological organisms is virtually dynamically incoherent with the organism/environment coupling, since the information needed to construct a given protein (the description) can be retrieved at any time as much as a book can be retrieved from a library[16]. Indeed, recent organisms carry many of the same genes used by primordial organisms to produce the same proteins. In their role of information carriers, the dynamical substrate of genes (the DNA molecule) is largely irrelevant.

Von Neumann showed that there is an advantage of symbolic, localized memory over purely dynamic, or distributed, memory in self-replication because if we do not have symbolic descriptions directing self-replication, then an organism must replicate through self-inspection of its parts. Clearly, as systems grow in complexity, self-inspection becomes more and more difficult[14]. The existence of a language, a symbol system, allows a much more sophisticated form of replication. Functional, dynamic structures do not need to replicate themselves, they are simply constructed from non-functional (dynamically inert) descriptions and available parts. For instance, for an enzyme to replicate itself without descriptions, it would need to have the intrinsic property of self-replication "by default" as a template, or it would have to be able to assemble itself from a pool of existing parts. But for this, it would have to "unfold" so that its internal portions could be reconstituted by self-inspection for the copy to be produced[14]. With the genetic code, however, none of these complicated gimmicks are necessary: functional molecules can be simply folded from inert messages. This method is by far more general since any functional molecule can be produced from a description, not merely those that either happen to be able to self-reproduce as a template, or those that can unfold and fold at will to be reproduced by self-inspection from available parts.

---

[v] For a detailed discussion of this model see[6, 7, 8].

The genetic symbol system, with its utilization of inert structures, opens up a whole new universe of behaviors and functionality which is not available to purely dynamic self-replication (template or self-inspection). In this sense, it can evolve behaviors and functions in an open-ended fashion: all describable (by the genetic code) proteins can be produced. It also introduces the third level of a semiosis of classifying systems in situation with their environments: syntax – as defined by a construction code. This type of system-environment selective coupling can be referred to as *description-based selected self-organization*[8]. Arguments for the idea of language as a provider of such an enabling syntax for cognitive systems which can be used to recombine distributed memory systems in an open-ended manner have been pursued elsewhere[17, 7]. In particular, a computer system for active recommendation of information in networked databases has been developed and proposed as a model of cognitive categorization[11]. In the cognitive realm, this type of system/environment coupling capable of open-ended semiosis is referred to as *linguistic-based selected self-organization*. Clearly, the evolutionary processes that result from this semiotic system-environment coupling at the biological and cognitive level are different, yielding active and passive evolutionary trends[18, 19, 20] which beg to be identified and modeled computationally[12].

### 3.2 Descriptions and Material Codes[vi]

Von Neumann used the notion of description-based universal construction clearly in a computer science context, where all construction must at some level be completely described or programmed. In his self-reproduction scheme a description encodes a complete blueprint of automata to be constructed by the universal constructor automaton. This need for a complete specification of the construction of self-reproducing automata is very far from the actual biological, material, machinery of the genetic selected self-organization of living systems. So in what sense do we use the term *description* when speaking of biological systems? Clearly, genes do not encode the complete specification of proteins and the means to produce them. That is, genes do not encode information such as how to fold a protein. All of this comes for free with the laws of matter[21]. A true universal constructor, in a physical sense, would need to encode <u>everything</u> down to physical law. Henri Atlan[22] has warned us about this computational metaphor, and proposes that instead of thinking of genes as descriptions or programs, we should think of them as data for the dynamic machinery of the cell. Let us explore this issue in more detail.

When Von Neumann's universal constructor interprets a description to construct some automaton, a *semiotic code* is utilized to map (translate) instructions into computational actions to be performed. When the copier copies a description, only its *syntactic* aspects are replicated (transcription). Now, the language of this code presupposes a set of computational primitives for which the instructions are said to "stand for"(the semantics). In other words, descriptions depend on the set of computational primitives of the interpreting device (the constructor). If such a set of primitives is complete in the Turing sense, then it establishes a <u>computationally</u> universal process.

---

[vi] The issue of what is meant by descriptions was raised by Alvaro Moreno at this meeting. While this issue is not fundamental to the argument pursued in this article, allowing this section to be bypassed without loss, it is nonetheless a very interesting topic. Thanks to Alvaro for pointing this out.

In contrast, in the material world, a construction code such as the genetic code presupposes a set of material primitives that follow physical law. In this sense, material codes are not universal as they refer to some material constituents which cannot be changed without altering the significance of the descriptions. We can see that a self-reproducing material organism following Von Neumann's scheme is an entanglement of *symbolic controls* and *material constraints* which is closed on its semantics only through its repercussions in an environment (the pragmatics). Howard Pattee calls such a principle of self-organization *semantic closure*[14, 23]. Perhaps a better description would be to refer to it as *semiotic closure* since then the three semiotic dimensions of semantics, pragmatics and syntax would be well accounted for.

A material code is defined by a small, finite, number of symbols (e.g. codons in DNA), which can encode a finite number of primitive parts (e.g. aminoacids). There is a countable number of functional structures which may be constructed with a given set of parts. This defines the representational power of a given material symbol system. From coded messages, a potentially infinite number of products can be constructed. However, since the products are dynamic and not symbolic structures, they will have different dynamic characteristics (for which they are ultimately selected). Moreover, the messages encoded stand for some arrangement of parts (strings of aminoacids) and not just the parts themselves. An arrangement of dynamic structures, however simple, tends to form a complex dynamic compound which will self-organize according to physical laws.

A particular material symbol system is tied to specific construction building blocks. The richer the parts, the smaller the required descriptions, but also the smaller the number of constructable products. Conrad[24] refers to this as a tradeoff between programmability and high evolutionary plasticity or efficient use of computational resources. For instance, Von Neumann used simple building blocks such as "and" and "or" gates to build the code of his self-reproducing automata, which in turn required a 29 state cellular automata lattice and very complicated descriptions. Arbib[25, 26] was able to simplify von Neumann's model greatly by utilizing more complicated logical building blocks, but losing some generality. Likewise, the genetic system does not need to describe all the chemical/dynamical characteristics of a "desired" protein (full programmability), it merely needs to specify an aminoacid chain which will itself self-organize (fold) into a functional configuration with some reactive properties. A given set of parts such as amino acids, provides intrinsic dynamic richness which does not have to be specified by the symbol system on which construction commands are cast[21] making descriptions much smaller and establishing higher evolutionary plasticity. The cost of this plasticity or efficient ability to specify proteins, is that the genetic code is not universal in that it cannot specify anything whatsoever, but only those things that can be constructed from aminoacid chains. It establishes nonetheless a material symbol system which can specify or describe <u>any</u> conceivable amino-acid chain. As Von Neumann showed, self-reproduction which utilizes such a symbol system, plus mutation, leads to open-ended evolution.

Interestingly, we can easily model computationally the ability to compress descriptions in symbol systems based on richer dynamic parts. That is, we can define computational codes which do not completely specify a product, but rather encode building blocks that can then self-organize computationally. This is often done in Artificial Life with indirectly encoded genetic algorithms: genetic descriptions which encode cellular automata, boolean networks, L-Systems and the like that self-organize into solutions interpreted at some higher level of observation of the simulation[27, 28]. In fact, I specifically showed how this kind of indirect encoding can solve optimization problems with smaller descriptions, as observed in material systems[7].

Therefore, we should think of descriptions not as complete, universal specification, but rather as material specification , that is: encoded initial conditions or constraints for material (e,g. aminoacid chains) or computational (e.g. cellular automata) self-organizing systems. This is how Pattee sees descriptions in his semantic closure principle: the symbolic controls that impose material constraints[14]. This seems better than Atlan's view of genetic information as data for the machinery of the cell. By trying to escape the overused computer metaphor, Atlan actually adds to the confusion of the computationalist metaphor by conceiving the machinery of the cell as a computer which needs data. The Pattee/Von Neumann view of genetic information as descriptions of initial conditions is more reasonable because the symbolic and material aspects of evolving systems are clearly delineated and not blurred: symbolic descriptions encode material constraints or initial conditions. This view of description-based selected self-organization treats descriptions not as programs but encoded self-organizing processes.

### 3.3 Why do we need syntax?

It can always be argued that the random access memory the genetic system establishes, is nothing but complicated dynamics, and the syntactic dimension is just the result of our subjective observation. But similar arguments can always be pursued to discourage any kind of multi-level theory or complementarity. Indeed, the notion of self-organization also requires a multi-level argument. The dynamic/self-organizing level results from the necessity of complementary modes of description to describe our (ultimately subjective) observation of patterns and regularities at another level other that state-transition rules. So why stop there? The genetic dimension has established a new hierarchical level in evolutionary systems which allows a greater level of control of the purely self-organizing bio-chemical dynamics. Failing to recognize this symbolic level, would prevent the distinction between self-organizing systems such as autocatalytic networks, from living systems whose replication by genetic memory is much more efficient than replication by template or self-inspection.

In evolutionary systems this is at the core of the feud between those who claim that natural selection is the sole explanation for evolution and those who stress that other aspects of evolutionary systems, such as developmental constraints, also play an important role. It is no wonder then that the first group stresses the symbolic description, the gene, as the sole driving force of evolution[29, 30], while the second group likes to think of the propensities of matter or historical contingencies as being of at least equal importance in evolution[31,32,9]. In pragmatic terms, however, most evolutionary theorists, acknowledge that all these factors play important roles[31].

Since all of these aspects of evolutionary systems co-exist, we need inclusive theories and models that incorporate both symbolic and dynamic characteristics[14,7,33]. We may conceive evolving agents which are purely dynamic; they observe the dynamically coherent selected self-organization discussed in 2 that is capable of semantic emergence in a selective environment (pragmatics). But the introduction of the syntactic level as prescribed by Von Neumann defines a richer (open-ended) evolution available only to agents capable of a full situated semiosis (semantics, pragmatics, and syntax) with their environments. In the next section we discuss how the former can be seen as autonomous agents in only a much weaker sense than the latter.

## 4. SYNTACTIC AUTONOMY

Semiotics leads us to think of symbols not simply as abstract memory tokens, but as material tools[34] for a situated semiosis of classifying systems with their environments, which requires the definition of components that interact and self-organize with the laws of their environment[2]. How such a semiotic code can arise from a purely dynamic self-organizing system is still very much a mystery both for biological and cognitive systems, though computational experiments to investigate the emergence of symbolic activity[35] and even codes[36] have been proposed. Before presenting some recent results in this area, we need to discuss what are the implications of the notion of semiotic (description or linguistic) selected self-organization for autonomous agents or agents participating in some closed loop with other agents or the environment.

A classifying self-organizing system is often described as autonomous if all processes that establish and sustain its dynamics are internally produced and re-produced over and over again[37]. These are the systems capable of *self-reference*[38]. But the autonomy or closure of these systems is really an abstraction that demands at some point an arbitrary identification of a boundary of some kind between system and environment. Indeed, the emergent, self-organizing, behavior of situated agents (dynamically coherent with their environments) is as much a result of the production rules of the agent as of the laws of the environment[39, 40, 2]. Furthermore, how autonomous are the systems that follow some form of situated semiosis with their environments? Given the arguments for selected self-organization, we know that it is the environment which ultimately selects the dynamic configurations of classifying systems. The dynamically coherent coupling between system and environment on which situated semiosis is based requires this pragmatic openness, *other-reference*[38, 41], or external scaffolding[40].

Semantics is therefore defined only by the situated, pragmatic, conjugation of system and environment, which indicates that situated agents are neither dynamically nor semantically autonomous. Indeed, a semantic loop or closure[23] can only be established between the agent and the entire environment, since, as we discussed above, situated semantics is pragmatic. If, for situated (dynamically coherent) agents, semantic closure can only be observed if we create a loop between an agent its entire environment, then we may be better off abandoning the notion of closure, or indeed semiosis, altogether and adopting a purely dynamical view of the world, in which agents are just another indistinguishable dynamical component of a network of many components. In this case, it might be much more accurate to use a dynamical systems theory explanatory scheme[39] [vii]. But is there any kind of more profound semiotic autonomy or closure in evolutionary systems that may require another level of explanation and modeling beyond dynamical systems theory?

Biological systems have developed a system of structural perturbation of their self-organization clearly based on a (genetic) code that essentially implements Von Neumann's scheme of inert symbolic descriptions. It is undeniable that this syntactic code is completely specified within organisms since its reading and constructing machinery is found within each cell: an autonomous code defined by specific syntactic rules. Even though environmental conditions clearly affect what is decoded in different circumstances[42, 7], the code itself remains fixed. The ability to generate such a powerful system of assembly of self-organizing encoded components for

---

[vii] In this paper I discuss autonomy and closure in terms of self-organization as a state-determined or dynamical process, and in terms of the semiosis of agents in an environment. Often autonomy and closure are also discussed in terms of thermodynamics. I understand that this may be a very important explanatory tool, but which, I believe, is nonetheless not yet theoretically well-developed enough to describe such out-of-equilibrium phenomena as what is necessary to explain living systems.

the construction of evolving classifying systems, is the one defining characteristic of all known life forms, which somehow produced this code or *autonomous syntax* for a more efficient situated semiosis with the environment.

Until we include syntax in the semiosis of situated agents in an environment as discussed in3, we cannot really speak of any kind of autonomy and the only closure we can identify entails the whole environment, which is clearly not interesting a type of closure. The introduction of a syntactic code allows the kind of recombination of dynamic descriptions used for construction of organisms which leads to open-ended evolution when included in a self-replication scheme as specified by Von Neumann. It furthermore allows the communication of these descriptions to systems which possess the same semiotic code. This way, more interesting types of closure between agents are enabled. Now, semiotic loops or closures can be established between an agent and other agents that share the same syntax without including the entire environment. Biological organisms all share the same genetic code, thus establishing a full semiotic closure with semantic, pragmatic, and syntactic dimensions which does not include non-biological agents. This is why we can use bacteria to produce human proteins or any other kind of transgenetic technology. We can find even tighter semiotic closures amongst organisms of the same species, which can communicate and recombine their genetic descriptions more effectively.

To recap, purely situated agents which exist in dynamical coherence with their environments, possess a week sense of agency in, or autonomy from, their environments. Their dynamics are intertwined with the environment and their semantics emerges from repercussions of their situated action in the environment. In this sense, there is also no physical or semiotic autonomy. Furthermore, a semiotic loop or closure can only be identified if we include the entire environment. It is only when we include the third dimension of semiosis, syntax, in the sense described in section 3 that we can identify more interesting types of autonomy and closure among groups of agents. The agents which possess this syntactic autonomy, rely on the existence of symbolic memory and possess therefore an element of dynamical incoherence with their environment (the strong sense of agency) in addition to the dynamic situation of their predecessors. The question now is how can symbols arise from dynamical substrates and how can we approach the study of this process.

## 5. SYNTACTIC AUTONOMY IN COMPUTATIONAL ENVIRONMENTS

### 5.1 Emergent Particle Computation

A very interesting problem that genetic algorithms (GA's) have been used successfully in, is the evolution of Cellular Automata (CA) rules for the solution of non-trivial tasks. Certain CA rules are capable of solving global tasks assigned to their lattices, even though their transition rules are local (each cell computes its next value given the current value of the cells in its immediate neighborhood). One such tasks is usually referred to as the *density task*: given a randomly initialized lattice configuration (IC), the CA should converge to a global state where all its cells are turned "ON" if there is a majority of "ON" cells in the IC, and to an all "OFF" state otherwise. This rule is not trivial because the local rules of the component cells do not have access to the entire lattice, but can only act on the state of their immediate neighborhood.

Crutchfield and Mitchell[35] used a GA to evolve the CA rules for such a task. The GA found a number of fairly interesting rules, but a few of the runs evolved very interesting rules (with high fitness) which create an intricate system of lattice communication. Basically, groups of adjacent cells propagate certain patterns across the lattice, which as they interact with other such patterns "decide" on the appropriate solutions for the lattice as a whole.

An intricate system of signaling patterns and its communication syntax has been identified, and can be said to establish the emergence of embedded-particle computation in evolved CA's[35, 43]. The emergent signals (or embedded particles) refer to the borders of the different patterns that develop in the space-time diagrams. If the areas inside these patterns are removed, their boundaries can be identified as a system of signals with a definite syntax, or emergent logic grammar. This syntax is based on a small number of signals, ", $, *, (, 0, and : , and a small number or rules such as: " + * 6 : , meaning that when signals " and * collide, the : signal results. Please refer to the references above for more details.

These experiments are very interesting because from the interaction of self-organization (CA's) and selection (GA) a very simple semantics emerges from the selective pragmatics of the GA: the CA rule either classifies its initial lattice configurations correctly or incorrectly. Now, most CA rules evolved with this set up show very simple space-time patterns: they try to solve the problem by block-expansion, that is, when large neighborhoods of either "ON" or "OFF" states exist in the initial configuration, they are expanded. These block expansion rules solved the task in typical dynamical fashion: by taking into account only local information.

Instead, the system of particle computation uses signals that are capable of integrating distant global information to solve the task. These CA rules rely on a system or grammar of signals (a code) used to communicate across the lattice and compute the answer to the task: a sign system that grants great selective advantage to the rules capable of developing it. The particle computation system truly introduces a qualitatively different way of solving the task: through the emergence of syntax, which allows certain rules to gain access to global lattice information. Obviously, such a system does not possess the rich self-reproduction scheme of Von Neumann, but it does show how the emergence of syntax grants simple dynamical systems the ability to move from trivial to non-trivial classification of their interaction with an environment.

## 5.2 Increasing Arbitrariness: Logical Tasks

The signals of the emergent particle computation system in CA's, even though being a small set of discrete entities, are not full-fledged symbols in the senses described in section 1, because they do not possess the degree of arbitrariness required of pure symbols: the syntax may be specific to the task solved. However, very similar signals and grammars can be evolved to solve different tasks, e.g. the synchronization task[43]. In other words, this class of CA's can develop similar signals to solve different problems.

To increase the arbitrariness of the emergent syntax of these rules, we can evolve rules that are good at solving several tasks. I have conducted some experiments to evolve CA rules with radius 3 which can solve both the density task and some related logical tasks. To implement logical tasks, we divide the CA lattice in two halves (the center cell is not used). The first half is interpreted as the first bit, and the second half as the second bit. A bit is "ON" if there is a majority of "ON" cells in its half, and "OFF" otherwise. Notice that since the boundary conditions of the lattice are periodic, this lattice has two boundaries between the two halves or bits. The cells on the neighborhood of these boundaries compute their values from cells in both halves, which in most cases makes the computation on these boundaries unreliable. However, since we are looking for global communication across the lattice, we expect the local errors at the boundaries not to be too relevant for the global computation, especially as lattices grow in size.

12We can now define such logical tasks as the AND and the OR task, according to the values of the bits. For the AND (OR) task, for all values of the bits the lattice should converge to an all "OFF" ("ON") state, except when both bits are "ON" ("OFF"). These tasks are both related to the density task because when the density of both halves is below (over) 0.5, both bits are "OFF" ("ON"), leading to a desired final lattice with all cells "OFF" ("ON"). They differ for the cases when the two halves of the lattice have opposing densities. In other words, these tasks should perform the density task in each half, and then integrate the results, with the AND (OR) task biased by "OFF" ("ON") information on either half.

Several rules were evolved with a GA whose initial population was composed of some of the best rules evolved so far for the density task, and whose fitness function was derived from presenting each rule with 100 different initial lattices, 50 to be analyzed by the density task, and the other 50 by either the AND or the OR task. The 50 rules to be presented to the density task have their density of "ON's" uniformly distributed over the unit interval (just as the experiments described in 3.1). The 50 rules presented to the AND (OR) task are biased to a uniform distribution of lattices leading to at least one bit "OFF" ("ON") 50% of the time, and both bits "ON" ("OFF") the other 50%. If we were to use an unbiased generation of lattices, only 25% of the time would the case of both bits "ON" ("OFF") be generated, making rules that always tend to "OFF" ("ON") always too favorable.

From these experiments, several rules were evolved that can solve both the density task and one of the logical tasks very well. Also, unlike the density task, the performance of the logical tasks often increases with the lattice size, probably because the boundary errors described earlier loose relevance in some cases as the density situation in each bit has a larger lattice to be resolved. I would expect this behavior to be a consequence of the velocity of the particles evolved, but such an analysis will be left for future research. A more detailed analysis of the particle computation systems of these recently evolved rules is forthcoming. Table I presents some of the rules evolved in hexadecimal format (each hexadecimal digit should be converted to 4 binary digits to obtain the CA rule; the left bit is the least significant one).

**Table I**: Unbiased performance (random generation of 100000 IC's) for the density, AND, and OR tasks, for CA lattices of dimension 149, 599, and 999. The first 4 rules are some of the rules fed to the initial population of the GA described above; the last 4 rules are some of the best rules evolved with this GA.

| Rule | $P_{dens}$ | | | $P_{AND}$ | | | $P_{OR}$ | | |
|---|---|---|---|---|---|---|---|---|---|
| | 149 | 599 | 999 | 149 | 599 | 999 | 149 | 599 | 999 |
| 0504058705000F77037755837BFFB77F [Crutchfield/Mitchell]] | .773 | .725 | .707 | .713 | .73 | .738 | .664 | .578 | .548 |
| 000F730F001FFF0F000FFF0F001FFF1F [Das Rule] | .823 | .777 | .763 | .68 | .684 | .68 | .733 | .686 | .675 |
| 0500550505500550555FF55FF55FF55FF [Koza rule] | .823 | .766 | .73 | .679 | .674 | .644 | .727 | .671 | .642 |
| 0760437B0700413507600F7F47F577FF [Jouille Rule] | .833 | .788 | .771 | .656 | .642 | .62 | .747 | .736 | .743 |
| 0057005D005F005D085FFF7F405FFF5F | .78 | .705 | .668 | .77 | .783 | .784 | .634 | .501 | .453 |
| 005F1053405F045F005FFD5F005DFF5F | .635 | .510 | .503 | .84 | .76 | .754 | .441 | .261 | .254 |
| 005F005F005F005F005FFF6F005FFF5F | .805 | .755 | .737 | .624 | .605 | .581 | .756 | .738 | .743 |
| 0504070705002573077755B37BFFF77F | .745 | .65 | .61 | .501 | .421 | .371 | .784 | .793 | .785 |

The relevance of these experiments is that they show that there is a family of particle computation rules which with a few mutations can develop a system of particle computation that can solve two different, yet related, tasks. Indeed, the rules were evolved from a population of rules that solve very well the density task. The particle computation systems provides the self-organizing CA the ability to adapt to a new environment that requires the solution of two similar tasks. In other words, it has the ability to evolve into a system that with the same syntax can effectively solve a related class of problems and not just one single task. In this case the class of tasks includes the density task and some logical task that is coherent with the density task. The ability to solve more than one task increases the arbitrariness of the emergent syntax of these rules, as the same syntactic rules of particle computation are used to compute different tasks. This increased arbitrariness shows that the particle-computation system can develop a larger scope of computations with particles that more and more be regarded as arbitrary symbols.

In fact, recent results from experiments conducted by Wim Hordijk[viii], show that equivalent sets of particles and grammars have been discovered to solve different tasks. This increases the scope of tasks that the same CA rule can solve, therefore granting higher arbitrariness and survival ability of their respective CA's when embedded in more demanding environments requiring different tasks to be solved. The CA's with particle computation ability, cope better with changing environments requiring different tasks, than purely dynamic CA strategies.

## 6. RNA Editing: the Origin of Syntactic Autonomy in Biology?

The discovery of messenger RNA (mRNA) molecules containing information not coded in DNA, first persuaded researchers in molecular biology that some mechanism in the cell might be responsible for post-transcriptional alteration of genetic information; this mechanism was called 'RNA Editing'[44]. "It was coined to illustrate that the alterations of the RNA sequence (i) occur in the protein-coding region and (ii) are most likely the result of a post-transcriptional event"[45, page 16]. The term is used to identify any mechanism which will produce mRNA molecules with information not specifically encoded in DNA. Initially, the term referred to the insertion or deletion of particular bases (e.g. uridine), or some sort of base conversion (e.g. adenosine ÷ guanisine). Today, more RNA editing mechanisms, have been observed, for a good review please refer to[46].

The idea that life may have originated from pure RNA world has been around for a while[47, 48]. In this scenario the first life forms relied on RNA molecules as both symbolic carriers of genetic information, and functional, catalytic molecules. The neutralist hypothesis for the function of RNA editing assumes such a RNA world origin of life. It posits that RNA editing could offer a process by which the dual role of RNA molecules as information carriers and catalysts could more easily co-exist. The key problem for the RNA world origin of life hypothesis is precisely the separation between these two functions of RNA. On the one hand RNA molecules should be

---

[viii] From recent experiments which will soon be published.

stable (non-reactive) to carry information, and on the other hand they should be reactive to perform their catalytic function. RNA editing, could be seen as means to fragment genetic information into several non-reactive molecules, that are later, through RNA editing processes, integrated into reactive molecules[46]. This way, the understanding of this process of mediation between the role of RNA molecules as information carriers and catalytic molecules based on RNA editing, can also offer many clues to the problem of origin of a semiotic code from s dynamic (catalytic) substrate.

Given many random distributions of the reactivity of a RNA sequence space, we could study how easily can reactive sequences be constructed from RNA edition of non-reactive molecules. A study of this process is forthcoming.

## 7. CONCLUSIONS

I discussed the notions of closure and autonomy of evolutionary agents in terms of self-organization as a state-determined or dynamical process, and in terms of the semiosis of such agents in an environment. Based on the work of Pattee and Von Neumann, I propose that until a private syntax (syntactic autonomy) is discovered by self-organizing agents, these agents exist in dynamically coherence or situation with their environments that include other agents. At this stage there are no significant or interesting types of closure or autonomy. When syntactic autonomy is enabled, then, because of description-based selected self-organization, open-ended evolution is established. At this stage, agents depend on localized, symbolic memory, thus adding a level of dynamical incoherence to their interaction with the environment. Furthermore, it is the appearance of syntactic autonomy which enables much more interesting types of closures amongst agents which share the same syntax.

Particle computation in cellular automata experiments were also discussed as an example of how we can study computationally the emergence of syntax from self-organizing dynamics. RNA editing is also mentioned as a process that may have been used to obtain the kind of syntactic autonomy necessary for open-ended evolution, that is, a description-construction code.